# Description and Evaluation of Semantic similarity Measures Approaches

Thabet Slimani
Computer Science Department
Taif University & LARODEC
Lab
thabet.slimani@gmail.com

## ABSTRACT

In recent years, semantic similarity measure has a great interest in Semantic Web and Natural Language Processing (NLP). Several similarity measures have been developed, being given the existence of a structured knowledge representation offered by ontologies and corpus which enable semantic interpretation of terms. Semantic similarity measures compute the similarity between concepts/terms included in knowledge sources in order to perform estimations. This paper discusses the existing semantic similarity methods based on structure, information content and feature approaches. Additionally, we present a critical evaluation of several categories of semantic similarity approaches based on two standard benchmarks. The aim of this paper is to give an efficient evaluation of all these measures which help researcher and practitioners to select the measure that best fit for their requirements.

## 1. INTRODUCTION

Semantic similarity between concepts is a method to measure the semantic similarity, or the semantic distance between two concepts according to a given ontology. In other terms, semantic similarity is used to identify concepts having common "characteristics". Although human do not know the formal definition of relatedness between concepts, he can judge relatedness between them. For example, a small child can tell that "apple" and "peach" have more related to each other than "apple" and "tomatoes". These pairs of concepts are related to each other and its structure definition is formally called "is-a" hierarchy. Semantic similarity methods becoming intensively used for most applications of intelligent knowledge-based and semantic information retrieval systems (identify an optimal match between query terms and documents) [1] [2], sense disambiguation [3] and Bioinformatics [4]. Semantic similarity and semantic relatedness [5] are two related words, but semantic similarity is more specific than relatedness and can be considered as a type of semantic relatedness. For example 'Student' and 'Professor' are the related terms, which are not similar. All the similar concepts are related and the vice versa is not always true.

Semantic similarity and semantic distance are defined conversely. Let be $C1$ and $C2$ two concepts that belong to two different nodes $n1$ and $n2$ in a given ontology, the distance between the nodes ($n1$ and $n2$) determines the similarity between these two concepts $C1$ and $C2$. Both n1 and n2 can be considered as an ontology (also called concept nodes) that contains a set of terms synonymous and consequently. Two terms are synonymous if they are in the same node and their semantic similarity is maximized.

The use of ontologies to represent the concepts or terms (humans or computers) characterizing different communicating sources are useful to make knowledge commonly understandable. Additionally, it is possible to use different ontologies to represent the concepts of each knowledge source. Subsequently, the mapping or concepts comparing based on the same or different ontologies ensures knowledge sharing between concepts. The mapping needs to find the similarity between the terms or concepts based on domain specific ontologies. The similarity between concepts or entities can be identified if they share common attributes or if they are linked to other semantically related entities in an ontology [6,7]. For example, the mapping between the KIMP ontology and MeSH ontology helps to identify the relationship with the standardized medical terms which improves the reusability and the discovery of the more related concepts.

This paper focus on semantic similarity. It enumerates four categories of semantic similarity measures described in literatures. Each approach of semantic similarity measure has been compared to others in the same category and evaluated.

The rest of this paper is structured as follows. Section 2 describes some examples of recognized ontologies used with semantic similarity measures. Section 3 presents the categories of semantic similarity measures. Section 4 gives an evaluation of the described semantic similarity measures. Section 5 is the conclusion.

## 2. EXAMPLES OF ONTOLOGIES USED WITH SEMANTIC SIMILARTY MEASURES

There are several examples of ontologies available including:WordNet [8][9], SENSUS[1] [10], Cyc[2][11], UMLS[3] [12], SNOMED[4], MeSH [13], GO[5] [14] and STDS[6]. The following section classify ontologies into general purpose ontologies and domain specific ontologies as follows:

### 2.1 General Purpose Ontologies

*2.1.1 Wordnet*

Wordnet is a lexical reference system developed at Princeton University with the attempt to model the lexical knowledge of a native speaker of English. It is an online database including nouns, verbs, adjectives and adverbs grouped into sets of cognitive synonyms (synsets), each expressing a distinct concept. Wordnet can be used to compute the similarity score and can be seen as an ontology for natural language terms. The latest online version of WordNet is v.3.1 announced in June 2011 and contains around 117,659 synsets and 206,941 word-sense pairs, organized into taxonomic hierarchies. Nouns, verbs, adjectives and adverbs are grouped into synonym sets (*synsets*). According to Wordnet, the synsets are also organized into synonym set corresponding to different synonyms of the same term or concept. Different types of relationships can be derived between the synsets or concepts (related to other synsets higher or lower in the hierarchy). The *Hyponym/Hypernym* relationship (i.e., Is-A relationship), and the *Meronym/Holonym relationship* (i.e., Part-Of relationship) are the most recognized relationships in WordNet. WordNet can be used as both a thesaurus and a dictionary. A fragment of the WordNet Is-A hierarchy is illustrated in Figure 1.

*2.1.2 SENSUS:*

SENSUS is an extension and reorganization of WordNet which contains a 90000 node concept thesaurus. The nodes adding is realized at the top level of the Penman Upper Model, additionally to the rearrangement of the major branches of WordNet. Each concept in SENSUS is represented by one node, i.e., each word has a unique specific sense, and the concepts are linked in an IS-A hierarchy.

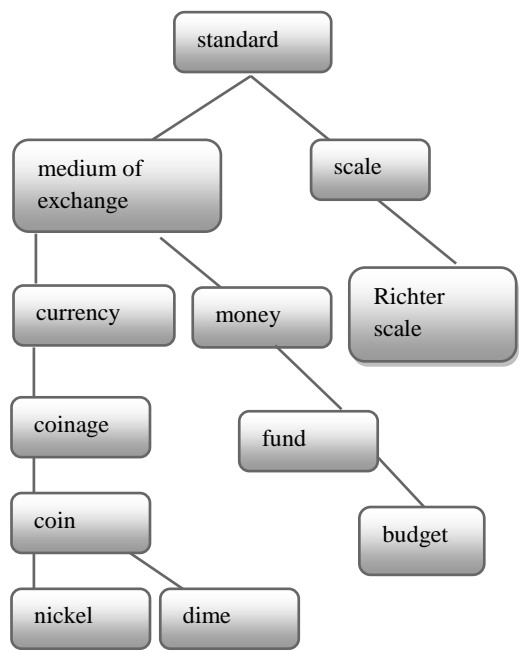

**Fig1: A fragment of the WordNet hypernym hierarchy**

*2.1.3 Cyc KB*

Cyc is a knowledge base designed to serve as an encyclopedic repository of all human knowledge primarily commonsense knowledge. Cyc is composed by terms and assertions relating those terms. As an example of fundamental human knowledge that can be included in Cyc: facts, rules of thumb, and heuristics for reasoning about the objects and events of everyday life. At the present time, the Cyc KB contains over five hundred thousand terms, including seventeen thousand types of relations, additionally to a seven million assertions which relates these terms.

### 2.2 Domain Specific Ontologies

*2.2.1 UMLS*

The Unified Medical Language System (UMLS) contains a very large, multi-purpose and multilingual metathesaurus containing information about biomedical and health related concepts. It is built from the electronic versions of some different thesauri, code sets, classifications, and lists of controlled terms. UMLS contains information about over 1 million biomedical concepts and 5 million concept names from more than 100 incorporated controlled vocabularies and classifications (some in multiple languages) systems. Each concept in the Metathesaurus is assigned to at least one "Semantic type" (a category), and certain "Semantic relationships" may obtain between members of the various Semantic types. UMLS consists of the following components:

- Metathesaurus: UMLS database, a collection of the controlled vocabularies of concepts and terms and their relationships;
- Semantic Network

---

[1] http://mozart.isi.edu:8003/sensus2/
[2] http://www.cyc.com/kb
[3] http://www.nlm.nih.gov/research/umls
[4] http://www.snomed.org
[5] http://www.geneontology.org
[6] http://mcmcweb.er.usgs.gov/sdts/

- SPECIALIST Lexicona

*2.2.2 SNOMED*

SNOMED is a dynamic, scientifically validated clinical health care terminology and infrastructure that makes health care knowledge more usable and accessible. As a terminology, it is agreed that SNOMED is the most complete, multilingual clinical healthcare in the world. Terms are attached to concept codes, which are themselves organized in a DAG. SNOMED provides a common language enabling a consistent way to capture, to share and to aggregate health data across specialties and sites of care. Clinical decision support, electronic medical records, disease surveillance, ICU monitoring, medical research studies, clinical trials, computerized physician order entry, image indexing and consumer health information services are among the applications for SNOMED.

*2.2.3 MeSH*

MeSH (Medical Subject Headings is a taxonomic hierarchy of medical and biological terms suggested by the U.S National Library of Medicine (NLM)[7]. It is organized as a set of terms naming *descriptors* in a hierarchical structure with more general terms (e.g " Body temperature changes ") higher in the taxonomy than most specific terms (e.g "fever"). There are 26,853 descriptors in 2013 MeSH[8].

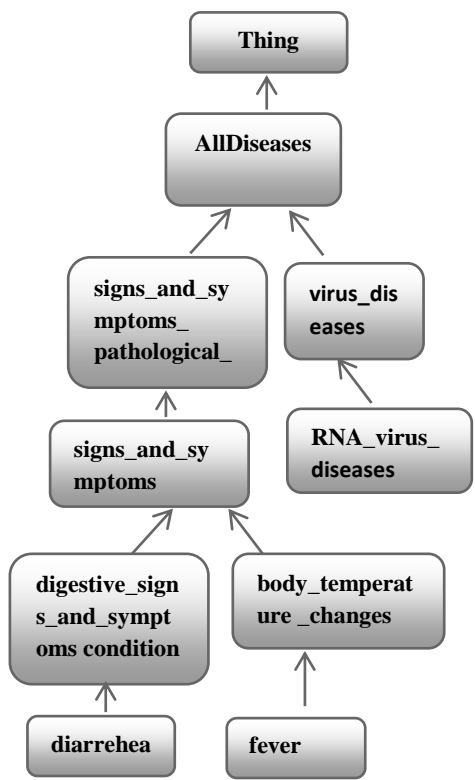

**Fig2: A fragment of the WordNet hypernym hierarchy.**

---

[7] http://www.nlm.nih.gov/

[8] http://www.nlm.nih.gov/pubs/factsheets/mesh.html

There are also over 213,000 entry terms that assist in finding the most appropriate MeSH Heading. There are more than 214,000 headings called Supplementary Concept Records within a separate thesaurus, in addition to these headings. A fragment of the WordNet Is-A hierarchy is illustrated in Figure 2.

*2.2.4 GO*

Gene Ontology (GO) describes gene proteins and all concerns of organisms as a structured network of defined terms. GO is developed based on a collaborative effort project to address the need for consistent descriptions of gene products in different databases. GO includes three structured controlled vocabularies (ontologies) that describe gene products in terms of their cellular components, associated biological processes, and molecular functions in a species-independent manner.

*2.2.5 STDS*

STDS (Spatial Data Transfer Standard) is a ***robust*** way of transferring earth-referenced spatial data between dissimilar computer systems with the potential for no information loss. It describes the underlying conceptual model and the detailed specifications for the content, structure, and format of spatial data, their associated attributes and features based on ontology. The commonly used concepts on topographic quadrangle maps and hydrographic charts are concepts in SDTS.

## 3. SEMANTIC MEASURE CATEGORIES

Several methods of determining semantic measures have been proposed in the last few decades. Three factors associated with the ontology taxonomic hierarchy can be specified: The path length factor, depth factor and local density factor in the hierarchy do affecting (although not significantly) the semantic distance measure. The density of two concepts C1 and C2 is the number of sons of the concepts which belong to the shortest path from the root to the most specific common subsumer of two concepts C1 and C2.

The similarity measures can affected by the common characteristics of the compared concepts. The differences between the concepts cause the measures to decrease or to increase with commonality. In addition, the similarity measures and the taxonomy can be related (taxonomic relations), i.e. the position of the concepts in the taxonomy and the number of hierarchic links are considered. Moreover, similarity measures take into account the information content of the concepts, whether they are enclosed or infinite values, whether they are symmetric and whether they give different perspectives. All the proprieties will be discussed in each class of similarity measure.

The proposed semantic measures are classified into four main classes or categories:

### 3.1 Structure-based measures

Structure-based or edge counting measures represent the measures that use a function that computes the semantic similarity measure in ontology hierarchy structure (is-a, part-of). The function computes the length of the path linking the terms and on the position of the terms in the taxonomy. Thus, the more similar two concepts are, the more links there are among the concepts and the more closely related they are [15] [16].

### 3.1.1 Shortest Path [15]

This measure is a variant of the *distance* method [15] and is principally designed to work with hierarchies. It it is a simple and powerful measure in hierarchical semantic nets. Let be C1 and C2 two concepts for which, the similarity measure in hierarchical structure can be formulated as follows:

$$Sim(C1, C2) = 2 * Max(C1, C2) - SP \quad (1)$$

Where Max is the maximum path length between C1 and C2 in the taxonomy and *SP* is the short path relating (minimum number of links) concepts C1 to concept *C*2.

### 3.1.2 Weighted Links [16]:

This measure is an extension of the above measure. It proposes weighted links to compute the similarity between two concepts. Two factors which affect the weight of a link: The depth of a specific hierarchy, the density of the taxonomy of a given level of the taxonomy and the strength of connotation[9] between parent and child nodes. Subsequently, the distance between two concepts is obtained by summing up the weights of the traversed links instead of counting them.

### 3.1.3 Hirst and St-Onge Measure (HSO) [17]

HSO measure calculates relatedness between concepts using the path distance between the concept nodes, number of changes in direction of the path connecting two concepts and the allowableness of the path. If there is a close relation between meanings of two concepts or words, then the concepts are said to be semantically related to each other [18]. An Allowable Path is a path that does not digress away from the meaning of the source concept and thus should be considered in the calculation of relatedness . Let be, *d* the number of changes of direction in the path that relates two concepts C1 and C2, and *C*, *k* are constants whose values are derived through experiments. The similarity function of HSO is formulated as follows:

$$Sim_{HSO}(C1, C2) = C - SP - k * d \quad (2)$$

### 3.1.4 Wu and Palmer [19]

Let be *C*1 and *C*2 two concepts in the taxonomy, this similarity measure considers the position of C1 and C2 to the position of the most specific common concept *C*. Several parents can be shared by C1 and C2 by multiple paths. The most specific common concept *is the* closest common ancestor *C* (the common parent related with the minimum number of IS-A links with concepts *C*1 and *C*2).

$$Sim_{wup}(C1, C2) = \frac{2*N}{N1+N2+2*N} \quad (3)$$

---

[9] The *connotation* of a term is the list of membership conditions for the denotation. The *denotation* of a term is the class of things to which the term correctly applies.

Where N1 and N2 are the distance (number of IS-A links) that separates, respectively, the concept C1 and C2 from the specific common concept and N is the distance which separates the closest common ancestor of C1 and C2 from the root node.

Consequently, the distance of the most specific common subsumer (N) has a non linear power. This evolution is observed as follows: if the sum is set to N1+N2 to a constant c: (power(∞) = ∞/(∞+c)). Furthermore, this measure is sensitive to the shortest path (c).

If it is required to calculate the Wu and Palmer similarity between *fev*er and diarrehea, i.e. *simwp*(*fever*, *diarrehea*) in Figure 1. This calculation is done as follows: firstly, determine that the least common subsumer of *fever* and *diarrehea* is signs_and_symptoms. Next, determine that the length of the path from *fever* to signs_and_symptoms is 2, that the length of the path from *diarrehea* to signs_and_symptoms is 2, and that the depth of signs_and_symptoms is 3. It is now straightforward to determine that *simwp*(*fever*, *diarrehea*)=$\frac{2*3}{2+2+2*3} = 0.6$.

### 3.1.5 Slimani et al. [20] (TBK):

Is an extension of Wu and Palmer measure with the attempt to improve edge counting results, because calculate the similarity of two terms in a hierarchy usually does not yield satisfactory results, in particular where this measure offers a higher similarity between a concept and its vicinity compared to this same concept and a concept contained in the same path. If a hierarchy Excerpt exists with the following format:

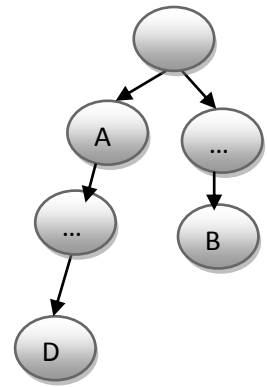

**Fig3: Example of hierarchy Exerpt.**

Applying the Wu and Palmer measure, it is possible obtain Simwp (A, D) < Simwp (A, B), where D is one descendant of A and B one of the descendants of the brothers of A. This situation is inadequate within the information retrieval framework where it is necessary to turn up all descendants of a concept (i.e request) before its neighborhood. To deal with this problem, the work in [20] proposes a penalization factor of two concepts C1 and C2 placed in the neighborhood to be multiplied by Wu and Palmer measure. This function aims to penalize or to reduce the value of similarity measure where two concepts are not in the same hierarchy.

### 3.1.6 Li et al. [21]

proposed to incorporate the semantic vector and word order to calculate sentence similarity. This similarity measure combines the shortest path length (SP) between two concepts C1 and C2, and the depth in the taxonomy (N) of the most specific common concept C, in a non-linear function.

$$Sim_{li}(C1, C2) = e^{-\alpha * SP} * \frac{e^{\beta * N} - e^{-\beta * N}}{e^{\beta * N} + e^{-\beta * N}} \quad (4)$$

Where $\alpha \geq 0$ and $\beta \geq 0$ are parameters scaling the contribution of shortest path length and depth respectively. The optimal parameters are $\alpha = 0.2$ and $\beta = 06$, based on [21]. It is therefore obvious that this measure scores between 1 (for similar concepts) and 0.

### 3.1.7 Leacock and Chodorow [22]:

The relatedness similarity measure proposed by Leacock and Chodorow (LC) is:

$$Sim_{LC}(C1, C2) = -\log\left(\frac{length}{2.D}\right) \quad (5)$$

Where length is the length of the shortest path between the two concepts (using node-counting) and D is the maximum depth of the taxonomy. Based on this measure, the shortest path between two concepts of the ontology restricted to taxonomic links is normalized by introducing a division by the double of the maximum hierarchy depth.

**Table 1. Structure-based semantic measures typology**

|  | Data Sources | Semantics | factors | | |
|---|---|---|---|---|---|
|  |  |  | SP | Concept Density | N |
| Shortest Path | Ontology | Distance | √ |  |  |
| Weighted Links | Ontology | Weighted links | √ | √ |  |
| Hirst and St-Onge | Ontology | Relatedness | √ |  |  |
| Wu and Palmer | Ontology | Similarity | √ | √ | √ |
| Slimani et al. | Ontology | Similarity | √ | √ | √ |
| Li et al. | Ontology | Similarity | √ |  | √ |
| Leacock and Chodorow | Ontology | Similarity | √ |  |  |

The previously described measures are based only on hierarchic (IS-A) links between concepts, taking into account that links in the hierarchy represent distances. Experimental results presented in [21] have demonstrated that the *Li et al.* measure has a good performance better than the previous proposed measures. Table 1 gives a comparing topology of the described approaches. The key properties of the structure-base measures presented in the previous sections are summarized in the Table 2.

**Table 2. Structure-based semantic key properties**

|  | [15] | [16] | [17] | [19] | [20] | [21] | [22] |
|---|---|---|---|---|---|---|---|
| Increase with commonality |  |  |  | √ | √ | √ | √ |
| decrease with difference | √ | √ | √ | √ | √ | √ | √ |
| position in hierarchy | √ | √ | √ | √ | √ | √ | √ |
| path length | √ | √ | √ | √ | √ | √ | √ |
| symmetric | √ | √ | √ | √ | √ | √ | √ |
| Different perspectives |  |  |  | √ | √ | √ | √ |

## 3.2 Information Content Measures

Information content (IC) based measures are those measures that use the information content of concepts to measure the semantic similarity between two concepts/terms. The information content value of a concept is calculated based on the frequency of the term in a given document collection. The next section presents a good number of semantic similarity measures. All of them use the information content of the shared parent of two terms *C*1 and *C*2 (Equation 6), where *S*(*C*1; *C*2) is the set of concepts that subsume *C*1 and *C*2. The two concepts can share parents by multiple paths. The minimum *p*(*C*) is used when there is more than one shared parent where *C is* the *most informative subsume* (MIS).

$$P_{mis}(C1, C2) = min_{C \in S(C1,C2)}\{p(C)\} \quad (6)$$

To calculate the similarity of two words, the information content of the most informative subsume is used.

### 3.2.1 Resnik [7]

This measure uses the information content of the shared parents. The principle of this measure is as follows: two concepts are more similar if they present a more shared information, and the information shared by two concepts C1 and c2 is indicated by the information content of the concepts that subsume them in the taxonomy. Resnik measure is formally defined as follows:

$$Sim_{Resnik}(C1, C2) = -ln(p_{mis}(C1,C2)) \quad (7)$$

This measure provides us with information such as the size of the corpus; a large corpus numerical value indicates a large corpus. The Resnik measure is considered somewhat coarse, since many different pairs of concepts may share the same least common subsumer.

### 3.2.2 Lord et al [23]:

In this work, the authors have studied the effect of using semantic similarity measures when querying DNA and protein sequence databases. They used Resnik's metric [7] to quantify semantic similarity between the terms in the GO DAG, which is the information content of the most informative subsumer of these terms;

### 3.2.3 Lin et al. [24]

The authors of this work have proposed a measure based on an ontology restricted to hierarchic links and a corpus. This similarity takes into account the information shared by two concepts like Resnik, but the difference between them is in the definition. The definition contains the same components as Resnik measure but the combination is not a difference but a ratio.

$$Sim_{Resnik}(C1,C2) = \frac{2*\ln((p_{mis}(C1,C2))}{\ln(p(c1))+\ln(p(c2))} \quad (8)$$

Hence, using this measure to compare the terms of an ontology presents a better ranking of similarity than the *Resnik* measure.

### 3.2.4 Jiang & Conrath [25]

In a similar manner as Resnik, the authors have used a corpus in addition to a hierarchic ontology (taxonomic links). The distance between two concepts C1 and C2, formulated in this work is the difference between the sum of the information content of the two concepts and the information content of their most informative subsumer:

$$Sim_{Jiang}(C1,C2) = 2*\ln p_{mis}(C1,C2) - (\ln P(C1) + \ln P(C2)) \quad (9)$$

This measure is insightful to the shortest path length between C1 and C2 and the density of concepts along this same path.

**Table 3. Information content semantic measures typology**

|  | Data Sources | Semantics | factors | | |
|---|---|---|---|---|---|
|  |  |  | SP | Concept Density | N |
| Resnik | Ontology+ Corpus | similarity |  | √ |  |
| Lord et al. | Ontology+ Corpus | similarity |  | √ |  |
| Lin et al. | ontology+ corpus | similarity | √ | √ | √ |
| Jiang & Conrath | Ontology+ Corpus | distance | √ |  | √ |

**Table 4. Information content semantic key properties**

|  | [7] | [23] | [24] | [25] |
|---|---|---|---|---|
| Increase with commonality | √ | √ | √ | √ |
| decrease with difference |  |  | √ | √ |
| Information Content | √ | √ | √ | √ |
| position in hierarchy | √ | √ | √ | √ |
| path length |  |  |  |  |
| symmetric | √ | √ | √ | √ |
| Different perspectives |  |  | √ | √ |

Table 3 gives a comparing topology of the described approaches. The key properties of the information content based measures presented in the previous sections are summarized in the Table 4.

## 3.3 Feature-Based Measures

The study of the features of a term is very important, because it contains valuable information concerning knowledge about the term. Feature based measure assumes that each term is described by a set of terms indicating its properties or features. The similarity measure between two terms is defined as a function of their properties (e.g., their definitions or "glosses" in WordNet) or based on their relationships to other similar terms in hierarchical structure.

### 3.3.1 Tversky [26]

The Tversky measure takes into account the features of terms to compute similarity between different concept, but the position of the terms in the taxonomy and the information content of the term is ignored. Each term should be described by a set of words indicating its features. Common features tend to increase the similarity and (conversely) non-common features tend to diminish the similarity of two concepts [26].

$$Sim_{tvsk}(C1,C2) = \frac{|C1 \cap C2|}{|C1 \cap C2|+\alpha|C1-C2|+(\alpha-1)|C2-C1|} \quad (10)$$

Where C1 and C2 represent the corresponding description sets of two terms t1 and respectively and $\alpha \in [0,1]$ is the relative importance of the non-common characteristics. The value of $\alpha$ increases with commonality and decreases with the difference between the two concepts. The determination of $\alpha$ is based on the observation that similarity is not necessarily a symmetric relation.

### 3.3.2 X-Similarity [27]

Petrakis et al., have proposed in 2006 a feature-based function called X-similarity which proposes a matching between words extracted from WordNet by parsing term definitions. Two terms are similar if the concepts of the words and the concepts in their neighborhoods (based on semantic relations) are lexically similar. Let be A and B two synsets or term description sets. Because not all the terms in the neighborhood of a term present a connection with the same relationship, set similarities are computed per semantic relationship (SR) type (e.g., Is-A and Part-Of). The proposed similarity measure is expressed as follows:

$$Sim_{xsim}(A,B) = \begin{cases} 1, if\ S_{synsets}(A,B) > 0 \\ max\{S_{neighb}(A,B), S_{descr}(A,B)\}, if\ S_{synsets}(A,B) = 0 \end{cases} \quad (11)$$

Let i a relationship type, the similarity for the semantic neighbors $S_{neighb}$ is formulated as follows:

$$S_{neight}(A,B) = max_{i \subset SR} \frac{|A_i \cap B_i|}{|A_i \cup B_i|} \quad 12$$

In a similar manner, if *A* and *B* denote the set of synsets or description for the term *a* and *b*, the similarity for descriptions $S_{descr}$ and synonyms $S_{synsets}$ are both computed as follows:

$$S(A,B) = \frac{|A \cap B|}{|A \cup B|} \qquad 13$$

### 3.3.3 Rodriguez et al. [28]

The proposed similarity measure can be used for single or cross ontology similarities. According to Rodriguez and Egenhofer, a concept is considered as an entity class. In this work, finding the similarity between the synonym sets of the entity classes, similarity between the distinguishing features of the entity classes and the similarity between semantic neighborhoods of the entity classes are used to identify the similarity between entity classes. The weighted aggregation of the similarity among the three specified components (synonym sets, features and neighborhoods) is the similarity function between entity classes. Consequently, the similarity between entity classes of the ontology p and of ontology q, is given as follows:

$$Sim_{Rod}(C1^p, C2^q) = W_w S_w(C1^p, C2^q) + W_u S_u(C1^p, C2^q) + W_n S_n(C1^p, C2^q)) \qquad (14)$$

Where $S_w$, $S_u$, and $S_n$ are respectively the measure of the similarity between synonym sets, features, and semantic neighborhoods among classes C1 of ontology p and classes C2 of ontology q and are calculated using Equation 10 (based on the Tversky feature-matching model.). $W_w$, $W_u$ and $W_n$ are the respective weights of the similarity of each specification component. Ww, Wu and Wn should be ≥0 and the sum of Ww, Wu and Wn should be equal to 1.

The similarity function $S_w$ is a word matching function used to determine the number of common words and different words in the synonym sets. The similarity function $S_u$ is a Feature matching function is used to find similarity between the distinguishing features of the entity class. And The similarity function $S_u$ is a similarity function that measure the similarity between semantic neighborhoods.

As a conclusion, the Knowledge Feature-based measures exploit more semantic than edge-counting approaches.

## 3.4 Hybrid Measures

Hybrid measures combine the structural characteristics described above (such as path length, depth and local density) and some of the above presented approaches. Although, their accuracy for a concrete situation is higher than more basic edge-counting measures, which depend on the empirical alteration of weights according to the ontology and input terms.

### 3.4.1 Knappe [29]

Knappe defines a similarity measure using the information of generalization and specification of two compared concepts. This measure is primarily based on the aspect that there may be *multiple paths* connecting two concepts. The proposed measure expression is formulated as follows:

$$Sim_{Knappe}(C1, C2) = p * \frac{|Ans(C1) \cap Ans(C2)|}{|Ans(C1)|} + (1-p) * \frac{|Ans(C1) \cap Ans(C2)|}{|Ans(C2)|} \qquad (15)$$

Where p is a value in [0,1] which determines the degree of influence of generalization. Ans(C1) and Ans(C2) correspond to description sets (the ancestor nodes) of terms C1 and $c2$ respectively. The reachable nodes shared by both $c1$ and $c2$ are $Ans(C1) \cap Ans(C2)$. In this similarity measure function, three major desirable properties are considered: (1) the cost of generalization should be 2 b) the cost for traversing edges should be lower when nodes are more specific and (3) further specialization implies reduced similarity.

### 3.4.2 Zhou et al. [30]

Zhou has proposed a measure taking into account information content based measures and path based measures as parameter. The proposed measure is expressed by the following formula:

$$Sim_{Zhou}(C1, C2) = 1 - k \left( \frac{\ln(len(C1,C2)+1)}{\ln(2*(deep_{max}-1))} \right) - (1-k) * ((IC(C1) + IC(C2) - 2 * IC(lso(C1,C2))/2) \qquad (16)$$

Where $lso(c, c)$ is the lowest super-ordinate of C1 and C2. From the previous formula it is noticed that, both IC and path have been considerate for the similarity measure calculation. The parameter k needs to be adapted manually for good performance. If k=1, formula 16 is path-based; if k=0, formula 16 is IC-based measure.

The key properties of the feature-based similarity measures and hybrid similarity measures presented in the previous sections are summarized in Table 5.

**Table 5. Feature-based and hybrid semantic key properties**

|  | [26] | [28] | [29] | [30] |
|---|---|---|---|---|
| Increase with commonality | √ | √ | √ | √ |
| decrease with difference | √ | √ |  |  |
| Information Content |  |  |  | √ |
| position in hierarchy |  |  | √ |  |
| path length |  |  | √ | √ |
| symmetric |  |  |  |  |
| Different perspectives | √ | √ | √ | √ |

## 4. Similarity Measures Evaluation

The evaluation of the accuracy of the similarity measures described above is not an easy process, given that the notion of similarity measure is a subjective human judgment. An objective evaluation should be based on existing benchmarks. Several authors created evaluation benchmarks including word pairs whose similarities were judged by humans. A first experiments have been conducted in 1965 [31] by Rubenstein and Goodenough, which regroups 51 native English speakers students. The authors judged 65 word pairs similarity selected from ordinary English nouns. The subjects were asked to rate them, from 0.0 to 4.0 scale, according to their "similarity of meaning". More recently and for a similar study, in 1991 Miller and Charles [32] have recreated the previous experiment based only on the subset of 30 noun pairs (taking 10 from the high level "between 3 and 4", 10 from the intermediate level "between 1 and 3", and 10 from the lower level " from 0 to 1") among the pairs of the original 65 pairs and its similarity was re-judged by 38 undergraduate students . The correlation compared to the experiment in the work [31] was 0.97. The same experiment has been repeated again by Resnik in 1995 [33], but including only 10 graduate students and post-doc researchers to judge similarity. The obtained correlation compared to the results in [32] was 0.96. As a comparison of the three results described above, pirro [34] have conducted a comparative study in 2009, based on 101 human subjects (both English and non-English native speakers). The obtained correlation by [34] is respectively 0.96 compared to [31], 0.95 compared to [32] and 0.97 compared to [33]. The obtained results show the existence of a high correlation between the results, although it is performed more than 40 years and with various sets of people, which indicates that the similarity measure is steady over the years indicating them a reliable source to compare measures.

**Table 6. Evaluation of Edge Counting, Information Content, Feature-based and Hybrid semantic similarity methods on WordNet.**

| Measure | Work were evaluated | Bench1 | Bench2 |
|---|---|---|---|
| Shortest Path | [27] | 0.59 | N/A |
| Weighted Links | [27] | 0.63 | N/A |
| Hirst and St-Onge | [36] | 0.78 | 0.81 |
| Wu and Palmer | [27] | 0.74 | N/A |
| Li et al. | [27] | 0.82 | N/A |
| Leacock and Chodorow | [27] | 0.82 | N/A |
| Leacock and Chodorow | [37] | 0.74 | 0.77 |
| Resnik | [27] | 0.79 | N/A |
| Resnik | [37] | 0.72 | 0.72 |
| Lord et al. | [27] | 0.79 | N/A |
| Lin et al. | [27] | 0.82 | N/A |
| Lin et al. | [37] corpus | 0.7 | 0.72 |
| Lin et al. | [30] intrinsic | 0.7373 | 0.7700 |
| Jiang & conrath | [37] | 0.73 | 0.75 |
| Jiang & conrath | [30] distance | -0.8218 | -0.8071 |
| Tversky | [27] | 0.73 | N/A |
| X-similarity | [27] | 0.74 | N/A |
| Rodriguez et al. | [27] | 0.71 | N/A |
| Zhou et al. | [30] | 0.8798 | 0.8702 |

It is useful to use the standard benchmarks and the correlation coefficient presented in [31] and [32] as a measure of evaluation which enables an objective comparison between measures (Bench1 indicates the benchmark of Miller and Charles, Bench2 indicates the benchmark of Rubenstein and Goodenough). The adopted correlation values are originally reported by [31] and [32] benchmarks and summarized them in Table 6, in order to evaluate the accuracy of related works. The results in Tab 6 indicate the following remarks: The lowest precision is presented by the shortest path length measure described in (rada et al.) [15] (0.59). This low precision refers to a path length that measure the connectivity of two concepts is not accurate to measure their specificity. The other edge-counting approaches also exploiting the relative depth of the taxonomy which offer an improved higher precision (0.74) are the approaches in [19] (wu and palmer) and [22] (Leacock and Chodorow). The result obtained with the work in [17] (Lin et al.) presents an improved result with regards to [15] this improvement refers to the use of non-taxonomic relationships that consider a more general concept of relatedness. The correlation value obtained by Li et al. [21] is more improved, because it combines the length of the path with the depth of the concepts in a weighted and non-linear manner.

Concerning IC-based measures, it is observed that intrinsic computation approaches which calculate the information content based on the number of concept hyponyms are clearly more accurate than corpora approaches (0.7373 vs. 0.70). This refers to the fact that corpora dependency seriously frustrate the applicability of classic IC measures.

Feature-based methods present a closer resemblance to those presented by structure-based measures (0.71-0.74). This refers to the fact that they rely on concept features (synsets, features or non-taxonomic relationships) which have secondary importance in ontologies and for that reason the approaches are based on partially modeled knowledge. As a consequence, those measures need more research to outperform the approaches based on edge-counting measures.

For hybrid-based measures, it is observed that the approach described in [30] offers the highest accuracy (0.87) even though it is a complex approach which exploits a relative depth and relying on weighting parameters.

## 5. CONCLUSION

Semantic similarity evaluation is a good factor included in many applications enclosed in the artificial intelligence research area. Based on the theoretical principles and the way in which ontologies are investigated to compute similarity, different kinds of methods can be identified. This paper provides an advanced examination of the most recognized semantic similarity measures that can be used to estimate the resemblance between concepts or terms. This paper has examined, with the aim of giving some insights on the accuracy, the typology and the key properties of the described measures under each category. In addition, an efficient comparison of all these measures in a practical setting is presented, using the two widely used benchmarks. The benefices concluded from those analyses would help the researcher and practitioners to select the measure that better fits with the requirements of a real application.